# Angle Based Feature Learning in GNN for 3D Object Detection using Point Cloud

Md Afzal Ansari[1][0000-0003-3588-702X], Md Meraz[2][0000-0003-4183-5779], Pavan Chakraborty[3][0000-0002-9260-1131] and Mohammed Javed[4][0000-0002-3019-7401]

[1,2,3,4]Indian Institute of Information Technology, Allahabad
[1]afzalansari880@gmail.com,{[2]pro2016001,[3]pavan,[4]javed}@iiita.ac.in

**Abstract.** In this paper, we present new feature encoding methods for Detection of 3D objects in point clouds. We used a graph neural network (GNN) for Detection of 3D objects namely cars, pedestrians, and cyclists. Feature encoding is one of the important steps in Detection of 3D objects. The dataset used is point cloud data which is irregular and unstructured and it needs to be encoded in such a way that ensures better feature encapsulation. Earlier works have used relative distance as one of the methods to encode the features. These methods are not resistant to rotation variance problems in Graph Neural Networks. We have included angular-based measures while performing feature encoding in graph neural networks. Along with that, we have performed a comparison between other methods like Absolute, Relative, Euclidean distances, and a combination of the Angle and Relative methods. The model is trained and evaluated on the subset of the KITTI object detection benchmark dataset under resource constraints. Our results demonstrate that a combination of angle measures and relative distance has performed better than other methods. In comparison to the baseline method(relative), it achieved better performance. We also performed time analysis of various feature encoding methods.

## 1      Introduction

In recent years deep learning-based methods solved many problems of computer vision like object classification [1-4], detection [5,6], and segmentation [2,3] in image data. But in real life, these algorithms can't apply to capture 3D environments for Autonomous vehicles. Because in an outdoor world a camera can't capture the surrounding information due to sunlight or rain and fog [7]. So, the new methods of surrounding capturing are performed using LiDAR and point cloud data. The point clouds are needs to pre-processed before applying to different domains. There are 2D detection techniques available but they don't capture the real world closely. Extension of 2D to 3D networks leads to bad performance and inaccurate results. 3D points cloud is irregular points of data (raw format)or irregular format of data so the informationretrieval is hard as well because 6-7 DoF (Degree of Freedom) [8] needs to be processed, which is challenging and computationally expensive [9]. Detection of 3D objects for self-driving cars requires the model to be accurate and time-efficient.

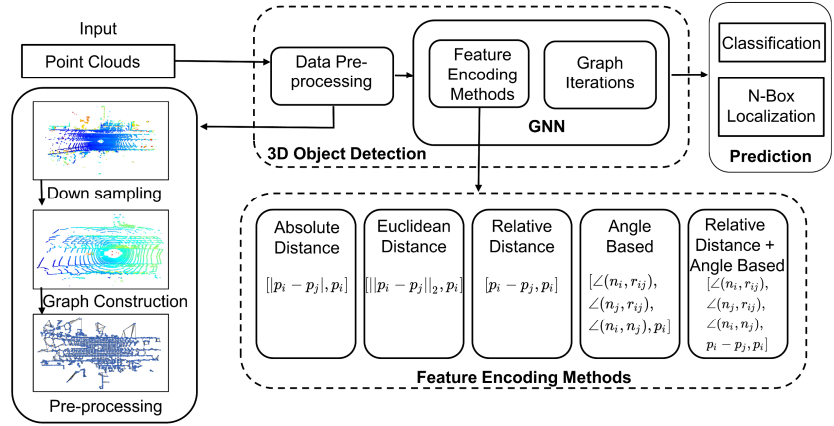

**Fig. 1.** Block Diagram shows the architecture of the Detection of 3D objects model with feature encoding methods. As shown in the diagram, five types of feature encoding are used in this work.

Point clouds are capable of modeling the real world in a better way as compared to images. But it has an unstructured and irregular data format. For object detection, either point cloud needs to be converted to a structured grid format i.e. voxels, views, RGB-D etc. [10-14, 5] or directly process the point clouds. Previous works have shown both the ways are promising ways of performing Detection of 3D objects. But converting the point cloud to other formats can lead to increased computation cost and loss of depth information.

Earlier works established Graph Neural Networks are potential neural networks for Detection of 3D objects in point cloud data [6]. GNN's requires input in form of graphs for their processing. During the formation of graphs, each point is modeled as a vertex and a feature vector is generated corresponding to each vertex through feature encoding. Here, Feature encoding means to encode the geometric orientation of a point w.r.t. the neighboring points. Relative distances between points is one of the methods for it. But due to less resistance to rotation variances, model accuracy drops when any rotational changes are introduced in the structure [15].

We proposed a new feature encoding method based on angle information learning of the point clouds. The angular information preserves the structure of the 3D scene during feature encoding of the point cloud. The addition of angular information of points along with relative distance better encodes the geometrical orientation and boost feature learning in GNN. Few works have been done in the field of feature encoding methods. Some of the prominent works includes [16] for feature encoding. Our works demonstrate the importance of feature encoding methods for ensuring accurate object detection in the point cloud.

In this work, the Graph Neural Network is used to perform Detection of 3D objects in Point Clouds. The model outputs the category of each of the vertex present in the graph and 7 DoF box coordinates of objects. We have evaluated the model on a sample of the KITTI benchmark dataset under limited resources on Google Colab. The GNN model with the new feature encoding method has performed better as compared to its baseline model [6]. In short, the main contributions to this work are:

- We proposed a new angle-based feature learning in GNN for Detection of 3D objects in Point Clouds.
- Our angle-based feature learning is rotational invariant hence useful in an unstructured dataset like point clouds.
- We trained and evaluated the GNN model with angle-based feature learning on the subset of the KITTI benchmark dataset and archive better performed compared to the baseline model [6].

## 2    Related Work

In the literature, few studies specifically focused on the Detection of 3D objects using Point Clouds using Graph Neural Networks. The prior methods can be classified based on pre-processing of point cloud for Detection of 3D objects. A few of the earlier works include point cloud conversion to regular grid format then applies 3D convolutions to detect objects like in images. One of the methods called MVCNN [17] consists of 3D object projection into views of multiple orientation and then view-wise feature extraction is performed. These features are then fused to perform object detection. MVCNN [17] forms a global vector using max-pooling of multi-view features. However, max-pooling in a particular view can lead to huge information loss as it retains the maximum element in a view. Previous methods also employ convolution-based operations by converting the point clouds to volumetric representation.

Daniel etal. [18] introduced a network that works on voxels representation of data called VoxNet for Detection of 3D objects. In volumetric representations, explicit conversion of irregular point clouds is required to voxel grids. Conversion of point cloud to regular grids can leads to uneven distribution of points in voxels and computation cost also increases.

Further networks are introduced which directly consume the point cloud using MLP based Networks where each point is independently processed using specific Multi-Layer Perceptron (MLPs) and then forwarded to a symmetric function is used to aggregate features into a global vector i.e. PointNet [3]. Such networks [3] have drawbacks because each point is considered independently hence its losses local information between points. Several other deep learning models were introduced like-PointNet++[2]. Graph-based networks are also achieved good results with better time complexity [6].

Some hybrid methods are also available which combines the advantages of combination of voxels and point cloud-based feature learning. PV-RCNN [19] is a two-stage network for Detection of 3D objects in point cloud data. In stage I the voxel to the key point scene encoding step is performed and in stage II the key points are transformed into grid features are extracted. The PointRCNN [5] is the two-stage network for Detection of 3D objects. The network also consists of step 1 3D proposals are generated in a bottom-up manner while in step 2 box proposals refining is performed w.r.t canonical coordinates. The research work Point to parts is also based on a 2 -stage process of Detection of 3D objects comprising of part-aware and part-aggregation stage [20]. The PVRCN++ model [21] an improvement of the previous model PVRCNN uses local vector representation for Detection of 3D objects.

Graph-based methods perform detection on point cloud by forming graphs. This work consists of graph neural network Point-GNN [6] in which graphs formed of points are taken as input. The network outputs the category and box localization of each vertex in the graph. The SVGA-net [22] is also one of the methods which consist of combined voxels and a graph-based approach. Feature encoding methods are needed in Graph Neural Networks. Earlier works show relative distance as one of the methods for feature learning [6]. The angle-based features are earlier used in PPF-net for Detection of 3D objects in 3D images [16].

Point-GNN is incapable of performing better in sparse points and it is not rotationally invariant. Our work uses the graph-based approach to perform object detection. It compares different feature encoding methods for better feature learning in GNN.

## 3  Methodology

This section describes about the implementation of Graph Neural Network and feature encoding method.

### 3.1  Methods

Graph based networks includes the following steps:

**GraphConstruction.** The pre-processing and construction of the graphs are performed from the point cloud dataset as shown in 1. One of the methods used to create the Graphs out of the point clouds comprises of threshold radial distance around the point. All the points are connected in that radial distance to form a graph. A graph G = (P, E) is constructed using radial distance as a measure. All the points which lie in a radial distance 'r' are connected through edges to form a graph. It is given by Eqn. (1):

$$ED = \{(p_i, p_j) \mid \left\|p_i - p_j\right\|_2 < r\} \quad (1)$$

where $p_i \in (x_i, y_i, z_i), p_j \in (x_j, y_j, z_j)$.

Before forming the graph, these points need to be down sampled as one frame consist of more than 200k points. Down sampling is performed using voxels. Here, we have used different feature encoding methods to embed features for further process.

---

**Algorithm 1** Algorithm for Angle based feature encoding
---

**Input :** P - PointCloud

**Output :** M – classification scores

0…M-1 Box Localizations

P ← [ $P_1, P_2, P_3 ... P_N$ ], Y ← [ 1, 2, 3 … N]

**Function** GetAngleFeatures (G, $K_p$):

> **for** $P_i$ in P **do**
>> **for** $P_j$ in $K_p$ **do**
>>> $$n_i = \frac{x_i + y_i + z_i}{\sqrt{x_i^2 + y_i^2 + z_i^2}}$$
>>>
>>> $$n_j = \frac{x_j + y_j + z_j}{\sqrt{x_j^2 + y_j^2 + z_j^2}}$$
>>>
>>> $$r_{ij} = \frac{(x_i - x_j) + (y_i - y_j) + (z_i - z_j)}{\sqrt[2]{(x_i - x_j)^2 + (y_i - y_j)^2 + (z_i - z_j)^2}}$$
>>>
>>> $$AM_1 = \arccos \frac{n_i . n_j}{|n_i||n_j|}$$
>>>
>>> $$AM_2 = \arccos \frac{r_{ij} . n_j}{|r_{ij}||n_j|}$$
>>>
>>> $$AM_3 = 180 - (AM_1 + AM_2)$$
>>>
>>> $$F_i = P_i, \{AM_1, AM_2, AM_3\}$$
>>
>> **end**
>
> **end**

**return** F

F = GetAngleFeatures (P)

---

**Feature Encoding**. The feature encoding method prepares a feature vector across each vertex. We have implemented and compared different feature encoding methods.

*Relative Distance.* Each of the points in point cloud data consists of (x, y, z, LiDAR reflectance intensity) as the parameters where (x,y,z) is the point cloud's cartesian coordinate. To form the feature vector, we have used relative distance as the metric. For a point $p_i$ it is given by Eqn. (2-3):

$$RD_i = \{(p_i, p_j) \mid (x_i - x_j, y_i - y_j, z_i - z_j)\} \tag{2}$$
$$FE_i = \{(p_i, p_j) \mid RD_i, Refl.Intensity)\} \tag{3}$$

where, $FE_i$ is Feature Encoded for a point $p_i$, $RD_i$ is Relative Distance between $(p_i, p_j)$

*Absolute Distance.* The feature vector is formed using absolute distance as the metric. For a point $p_i$ it is given by Eqn. (4-5):

$$AD_i = \{(p_i, p_j) \mid (||x_i - x_j||, ||y_i - y_j||, ||z_i - z_j||)\} \tag{4}$$
$$FE_i = \{(p_i, p_j) \mid AD_i, Refl.Intensity)\} \tag{5}$$

where $FE_i$ is Feature Encoded for a point $p_i$, $AD_i$ is Absolute Distance between $(p_i, p_j)$

*Euclidean Distance.* The feature vector is formed using Euclidean distance as the metric. For a point $p_i$ it is given by Eqn. (6-7):

$$ED_i = \{(p_i, p_j) \mid ((x_i - x_j)^2, (y_i - y_j)^2, (z_i - z_j)^2)\} \tag{6}$$
$$FE_i = \{(p_i, p_j) \mid ED_i, Refl.Intensity)\} \tag{7}$$

where $FE_i$ is Feature Encoded for a point $p_i$, $ED_i$ is Euclidean Distance between $(p_i, p_j)$

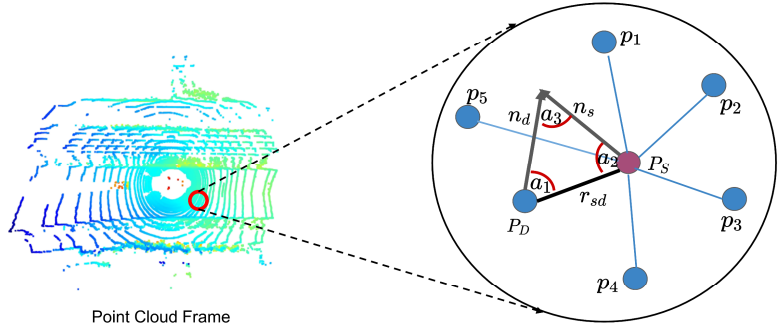

Point Cloud Frame

**Fig. 2.** In the above diagram angle-based feature encoding is shown. The magnified part shows a cluster of points where angle-based feature learning is performed. The feature vector is formed by the angles between normal vector $n_s$ of point $p_s$ and normal vector $n_d$ of point $p_d$, $n_s$ and relative distance $r_{sd}$, and $n_d$ and relative distance $r_{sd}$.

*Angular Measures.* Similarly, to form the angle-based feature vector, we have calculated angles between the points as shown in [angle]. We have formed two normal vectors corresponding to $P_i$, $P_j$ i.e $n_i$, $n_j$. Then we calculated a relative vector from $(P_i, P_j)$. A triangular region is formed which is then used to calculate the inner angles. The algorithm for the angle-based feature encoding is given in Algorithm 1. For a point $p_i$ it is given by Eqn. (8-9):

$$AM_i = \{(p_i, p_j) \mid (\angle(n_i, n_j), \angle(n_j, RD), \angle(RD, n_i)\} \tag{8}$$
$$FE_i = \{(p_i, p_j) \mid AM_i, Refl. Intensity)\} \tag{9}$$

where $FE_i$ is Feature Encoded for a point $p_i$, $AM_i$ is Angle measures.

*Combination of Relative Distance and Angular Measures.* The feature vector is formed by calculating the angle between two points. We have formed two normal vectors corresponding to $P_i$, $P_j$ i.e $n_i$, $n_j$. Then we calculated a relative vector from $(P_i, P_j)$. A triangular region is formed which is then used to calculate the inner angles. Then we combined it with relative distance to form a new feature vector. For a point $p_i$ it is given by Eqn. (10-12):

$$AM_i = \{(p_i, p_j) \mid (\angle(n_i, n_j), \angle(n_j, RD), \angle(RD, n_i)\} \tag{10}$$
$$RD_i = \{(p_i, p_j) \mid (x_i - x_j, y_i - y_j, z_i - z_j)\} \tag{11}$$
$$FE_i = \{(p_i, p_j) \mid (AM, RD, Refl. Intensity)\} \tag{12}$$

where $FE_i$ is Feature Encoded for a point $p_i$, $RD_i$ is Relative Distance between $(p_i, p_j)$, $AM_i$ is the Angle Measures.

**GNN architecture.** The architecture of the GNN is shown in Fig. 3. It takes input in form of point clouds, then a graph is constructed from it. The graph is passed through the pointset pooling function where initial vertex features are generated. Each vertex in the graph consists of a feature vector. A convolution MLP operation is performed as given in Eqn. 19. Further, all the vertex features are passed through max-pooling operation to form a single feature vector corresponding to each vertex. In the end, the vectors are passed through a predictor MLP of size (64,128,64) to obtain the M classification scores and M boxes localizations.

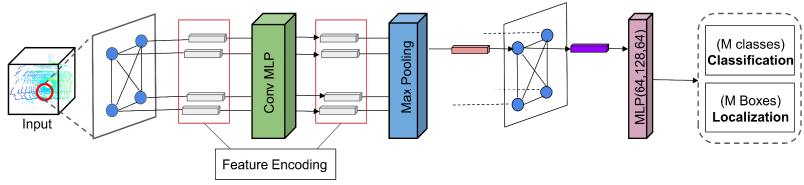

**Fig. 3.** The diagram shows the architecture of the GNN model with new feature encoding method.

**GNN iterations.** The GNN performs convolution through neural messaging approach. In case of GNN implementation for Detection of 3D objects, each vertex should include the feature vector which contain the geometric and relational information. To update the vertex's state by including the states of neighbor points:

$$v_i^{\{t+1\}} = g^t\big(\rho(\{fun^t((x_j - x_i), st_j^t)|(i,j) \in Edges\}), st_i^t\big) \qquad (13)$$

The state features of the vertex include relative coordinates that restricts the global change in the point cloud structure which is called as translation invariance property.

To deal with local translation invariances, an alignment offset is added to the relative coordinates. It can be given as follows:

$$\Delta x_i^t = h^t(st_i^t) \qquad (14)$$

$$v_i^{\{t+1\}} = g^t\big(\rho(\{fun^t(x_j - x_i + \Delta x_i^t, st_j^t)|(i,j) \in Edges\}), st_i^t\big) \qquad (15)$$

$\Delta x_i^t$ is the transitional invariance operand for each vertex which registers its coordinates.

A single iteration of the GNN can be given as follows:

$$\Delta x_i^t = MLP_h^t(st_i^t) \qquad (16)$$

$$e_{\{ij\}}^t = MLP_{fun}^t([x_j - x_i + \Delta x_i^t, st_j^t]) \qquad (17)$$

$$v_i^{\{t+1\}} = MLP_g^t(Max(\{e_{\{ij\}}^t|(i,j) \in Edges\})) + st_i^t \qquad (18)$$

The GNN at the end of T iterations, predicts the category of each of the vertex and the box parameters of each of the objects in which the vertex lies.

**Prediction of Class and Bounding Boxes.** Each of the vertexes of the GNN predicts a Bounding Box. Due to that multiple Bounding boxes are predicted. There is a need to merge all these boxes and assign a confidence score. The selection of an accurate bounding box can be done by considering an occlusion factor and classification scores by the Intersection-Of-Union factor. In the end, it returned a merged bounding Box.

### 3.2 Dataset

The point cloud is a group of points that are captured using LiDARs. Each of the point has (x, y, z) cartesian coordinates and additional information like color or laser reflectance intensity. It records the values like x, y, z coordinates, angle, intensity, and color of the surface. Similarly, KITTI Detection of 3D objects dataset comprises of point cloud dataset of objects like cyclists, pedestrians, cars with various levels of complexity that depend on easy, medium, and hard.The level of complexities varies depending on the coverage of the object in a particular point cloud. The KITTI benchmark detection dataset consists of 7481 training images and 7518 test images as well as the corresponding point clouds in form of Velodyne files. It comprises 80.256 labeled objects. We have used a subset of the KITTI dataset to train and evaluate the GNN model due to resource constraints. A dataset size of 1500 Velodyne point clouds is used for the same.

## 4    Experiments

The model is trained in terms of different Graph iterations. To deal with intensive memory consumption, we restricted the number of edges per vertex to 256. The feature vectors are generated using MLP (32,64,128,300). The $MLP_{class}$ is of size (64, #(classes)) and $MLP_{loc}$ is of size (64,64,7).

**Car**. As per the KITTI dataset we set the median of the Car bounding boxes to (3.88m, 1.5m, 1.63m). There are two views used for car detection. The front view of car varies from $\theta \in [\pi/4, 3\pi/4]$ and side view of car varies from $\theta \in [-\pi/4, \pi/4]$. In total it forms 4 classes along with DoNotCare class and Background class.

**Cyclist and Pedestrians**. We set the median of the Pedestrians bounding boxes to (0.89m, 1.76m, 0.64m). For the cyclist the median is set to (1.77m, 1.74m, 0.60m). There are two views used for cyclist and pedestrian detection as in the case of car detection. The front view varies from $\theta \in [\pi/4, 3\pi/4]$ and side view varies from $\theta \in [-\pi/4, \pi/4]$. In total it forms 6 classes along with DoNotCare class and Background class.

The model is trained with batch size =1. The SGD with stair-case learning parameter decay is used as optimizer for training GNN.The model is trained and evaluated on different feature encoding methods. It is trained on Google Colab under limited resources. The GPU provided by Google Colab is Tesla K80 GPU for 12 hours.

## 5    Results

We have evaluated and reported our results on a sample of the KITTI benchmark dataset. In Table 1. we compared the results on existing feature methods and our feature encoding methods. The metric used for comparison is mean Average Precision(mAP).

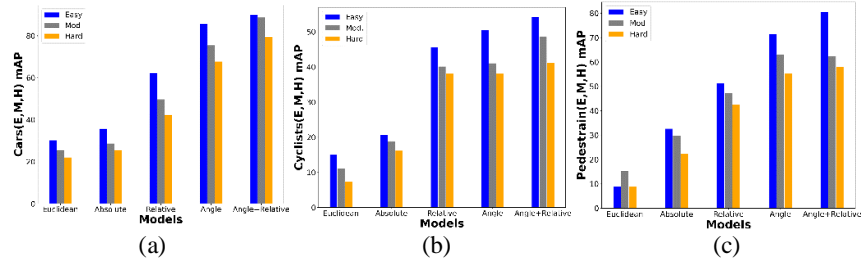

(a) (b) (c)

**Fig. 4.** The above bar-graphs shows mAP of the GNN detection model using the various feature encoding methods. (a) mAP of cars detection (b) mAP of Cyclist detection (c) mAP of Pedestrians detection

Based on the result the model with feature encoding method comprising of relative distance, and Angular measures achieves the best result. It is also clearly observable from fig. 4. (a-c) the result obtained by GNN model with angle-based feature encoding and combination of angle + relative based method has outperformed all other feature encoding methods by reasonable margin. The Model achieves the best result (90.12, 88.86, 79.53) mAP in Car detection for (easy, moderate, hard) while in the case of Cyclist the result achieved is (54.23, 48.67, 71.21) mAP, and in Pedestrians, the mAP achieved is (80.61, 62.41, 58.01).

**Table 1.** The table shows the results obtained on sample KITTI dataset using different feature encoding methods. The results reported in mAP for each of the objects under observation. Here E, M, H stands for Easy, Moderate and Hard respectively.

| GNN | Car(mAP) | | | Cyclist(mAP) | | | Pedestrians(mAP) | | |
|---|---|---|---|---|---|---|---|---|---|
| *Feature Encodings* | *E* | *M* | *H* | *E* | *M* | *H* | *E* | *M* | *H* |
| Euclidean | 30.23 | 25.58 | 22.02 | 15.15 | 11.06 | 7.47 | 8.93 | 15.47 | 8.93 |
| Absolute | 35.56 | 28.66 | 25.41 | 20.71 | 18.92 | 16.17 | 32.67 | 29.91 | 22.41 |
| Relative | 62.23 | 49.57 | 42.42 | 45.64 | 40.14 | 38.19 | 51.28 | 47.42 | 42.56 |
| Angle | 85.64 | 75.66 | 67.69 | 50.55 | 41.07 | 38.27 | 71.42 | **63.12** | 55.47 |
| Angle + Relative | **90.12** | **88.86** | **79.53** | **54.23** | **48.67** | **41.21** | **80.61** | 62.41 | **58.01** |

The model with the new feature encoding method surpassed the baseline method (Relative Method) by (27.89, 39.29, 37.11) in car detection on sample KITTI dataset. Along with the new feature encoding method, it also performed better in the case of cyclists detection by a margin of (8.59, 8.53, 3.02) but, in the case of pedestrians due to its sparse nature, it performed reasonably well. The Angle-based feature encoding also performed better than the baseline model (Relative method), but the combination of relative and angle has outperformed other methods.

## 6 Analysis

The time complexity analysis is shown in Table.2. The time complexity is reported in seconds elapsed during the various operations namely Generate graph, Inference, and

Table 2. The table shows the time analysis of the Detection of 3D objects using GNN model with different feature encoding methods.

| GNN | Time Complexity(sec) | | |
|---|---|---|---|
| *Feature Encoding* | *Gen. Graph* | *Inference* | *Total Time* |
| Euclidean | 0.088 | 0.567 | 0.750 |
| Absolute | 0.089 | **0.564** | 0.778 |
| Relative | **0.087** | 0.567 | **0.686** |
| Angle | 0.088 | 0.567 | 0.728 |
| Angle and Relative | 0.088 | 0.566 | 0.748 |

Total time. The Generate graph method is the construction of graphs from point cloud using feature encoding methods while the Inference time is the time required to fetch the result of detected objects in terms of metric(mAP). Based on the results reported the Relative distance method consumes less time as compared to other methods with a margin of three decimal places. The total time consumed by the Angle-Based method is 0.042 secs more than the baseline model (Relative Method) which is marginally small in comparison to the accuracy achieved by it.

## 7 Conclusions

In this work, we proposed a new angle-based feature encoding method for Detection of 3D objects from raw point cloud data. We showed that angle-based feature encoding is a potential feature encoding method for better feature learning in Graph Neural Networks. We have performed a comparison of different encoding methods with our angle-based feature encoding method and shown that it is better than the existing methods based on results. In the Detection of 3D objects problem, the combination of Angle-based and Relative distance-based feature learning has outperformed other methods evaluated on mean average precision(mAP). Using Angle-based feature learning the mAP improved by (27.89, 39.29, 37.11) in cars detection on sample KITTI dataset as compared to the baseline model. Similarly, it has performed well in the case of cyclists and pedestrian's detection. In future work, we will focus on modifying the down-sampling methods for raw point cloud data to retain more geometric information.